\documentclass[lettersize,journal]{IEEEtran}
\usepackage{amsmath,amsfonts}
\usepackage{algorithmic}
\usepackage{algorithm}
\usepackage{array}
\usepackage[caption=false,font=normalsize,labelfont=sf,textfont=sf]{subfig}
\usepackage{textcomp}
\usepackage{stfloats}
\usepackage{url}
\usepackage{verbatim}
\usepackage{graphicx}
\usepackage{cite}
\usepackage{booktabs}
\usepackage[normalem]{ulem}
\usepackage{dutchcal}
\usepackage{framed,multirow}
\usepackage{lineno,hyperref}
\usepackage{amssymb}
\usepackage{latexsym}
\usepackage{tabularx}
\begin{document}

\title{FAGStyle: Feature Augmentation on Geodesic Surface for Zero-shot Text-guided Diffusion Image Style Transfer}

\author{Yuexing~Han,~Liheng~Ruan,~and~Bing~Wang%
\thanks{Y. Han is with the School of Computer Engineering and Science, Shanghai University, 99 Shangda Road, Shanghai 200444, China, and Key Laboratory of Silicate Cultural Relics Conservation (Shanghai University), Ministry of Education (e-mail: Han\_yx@i.shu.edu.cn).}
\thanks{L. Ruan and B. Wang are with the School of Computer Engineering and Science, Shanghai University, 99 Shangda Road, Shanghai 200444, China.}}

\maketitle

\begin{abstract}
  The goal of image style transfer is to render an image guided by a style reference while maintaining the original content. Existing image-guided methods rely on specific style reference images, restricting their wider application and potentially compromising result quality. As a flexible alternative, text-guided methods allow users to describe the desired style using text prompts. Despite their versatility, these methods often struggle with maintaining style consistency, reflecting the described style accurately, and preserving the content of the target image. To address these challenges, we introduce FAGStyle, a zero-shot text-guided diffusion image style transfer method. Our approach enhances inter-patch information interaction by incorporating the Sliding Window Crop technique and Feature Augmentation on Geodesic Surface into our style control loss. Furthermore, we integrate a Pre-Shape self-correlation consistency loss to ensure content consistency. FAGStyle demonstrates superior performance over existing methods, consistently achieving stylization that retains the semantic content of the source image. Experimental results confirms the efficacy of FAGStyle across a diverse range of source contents and styles, both imagined and common.
\end{abstract}

\begin{IEEEkeywords}
  Image Style Transfer, Diffusion Models, The Shape Space Theory
\end{IEEEkeywords}

\section{Introduction}
\label{sec:Introduction}

Image style transfer represents a significant field in computer vision, where a specific style is applied to a source image while aiming to preserve the original content as accurately as possible. In this context, 'content' refers to the various instances in the source images, including its semantic information, outlines, structure and color, etc. For a long period, style transfer has depended on style reference images. Image-guided methods can transform a source image to match the style of iconic artworks. However, their reliance on specific reference images can restrict their broader application and potentially degrade the quality of the results. Additionally, most existing image-guided methods necessitate extra model training or fine-tuning, which complicates the process and diminishes efficiency. Finding suitable style images is also a labor-intensive process, especially when users wish to transfer a style for which no exact image exists.

Recent advances have led to notable progress in zero-shot text-guided style transfer, facilitated by the evolution of large pre-trained language models \cite{vaswani2017attention,openai2024gpt4technicalreport} and models that merge textual and visual data understanding \cite{radford2021learning,li2022blip}. The breakthroughs allow models to interpret and realize styles described only by text, introducing methods such as StyleCLIP \cite{Patashnik_2021_ICCV}, StyleGAN-NADA \cite{gal2022stylegan}, VQGAN-CLIP \cite{crowson2022vqgan}, and CLIPstyler \cite{kwon2022clipstyler}. Text-guided methods, employing either Generative Adversarial Networks (GANs) or encoder-decoder architectures, enhance style transfer capabilities by converting textual descriptions into visual styles.

Diffusion models (DMs) are also becoming increasingly popular for their exceptional efficacy in image generation and manipulation \cite{ho2020denoising,dhariwal2021diffusion,saharia2022photorealistic,ramesh2022hierarchical,rombach2022high,xu2023infedit,podell2024sdxl}. DMs have been adapted for both image-guided \cite{su2022dual,wang2023stylediffusion,zhang2023inversion,cho2024one,chung2024style,deng2024z} and text-guided \cite{song2021denoising,rombach2022high,kwon2022diffusion,yang2023zero,he2024freestyle} style transfer, yet they continue to face challenges in maintaining style consistency, accurately reflecting the described style, and preserving content in the target image.

To address these challenges, our study introduces a novel method termed Feature Augmentation on Geodesic Surface (FAGS) for zero-shot text-guided diffusion image style transfer, which we call FAGStyle. In the zero-shot scenario, a pre-trained image generation diffusion model, trained exclusively on unstyled, real-world photographs, is employed. Our objective is to seamlessly integrate style information from text prompts during the stylization of the source image, enabling the generation of stylized images without the need for additional training or fine-tuning. 

FAGStyle employs a sliding window crop technique to partition both the input source image and the intermediate generated target image into multiple patches. Features extracted from these patches are then projected into the Pre-Shape Space to construct a Geodesic surface. Similarly, features from the style reference text are projected into the same Pre-Shape Space to build the other Geodesic surface, facilitating effective style transfer by comparing the differences between the augmented image features and the text features from the corresponding Geodesic surfaces. The augmented features obtained through FAGS encapsulate information from each image patch, enhancing interactivity between these elements. The facilitated information interaction ensures that the stylization results produced through FAGStyle not only maintains a consistent and uniform style but also aligns closely with the style described in the text.

Moreover, we incorporate a Pre-Shape self-correlation consistency loss to further enhance the ability of our method to preserve the source content. Our experimental results demonstrate that our proposed method can successfully perform style transfer across a variety of styles and source input images from different domain while preserving content, indicating significant advantages over existing image-guided and text-guided methods.

\section{Related Work}
\label{sec:Related Work}

\subsection{Image Style Transfer}
\label{sec:Image Style Transfer}

In the fields of image processing and computer vision, style transfer has emerged as a pivotal technique that has evolved significantly over time. Initially, research efforts concentrated on manually modeling style images based on their texture statistics, aiming to match generated images to these predefined style constraints \cite{portilla2000parametric}. With the rise of deep learning, style transfer has undergone significant evolution, particularly with the development of Neural Style Transfer (NST) methods.

Generative Adversarial Networks (GANs) \cite{goodfellow2014generative} have been rapidly adopted for style transfer tasks due to their robust capabilities in generating high-quality images. CycleGAN \cite{zhu2017unpaired} facilitates style transfer without the need for pairwise samples by employing cyclic consistency loss and adversarial loss, although this method encounters certain limitations. Building on this, Contrastive Unpaired Translation (CUT) \cite{park2020contrastive} preserves structural consistency in content while maximizing mutual information across patches between the input and output images.

The integration of pre-trained multimodal models like CLIP \cite{radford2021learning} has further advanced text-driven image manipulation techniques. Methods such as StyleCLIP \cite{Patashnik_2021_ICCV} and StyleGAN-NADA \cite{gal2022stylegan} use the pre-trained CLIP to align latent vector of text with the style space of StyleGAN, enabling text-driven image editing. However, these methods remain constrained by the output domain of generative models. VQGAN-CLIP \cite{crowson2022vqgan} addresses this limitation by exploring innovative fusion techniques that combine generative models with CLIP text features, thus broadening the scope of image manipulation capabilities.

On the other hand, encoder-decoder architectures employing Convolutional Neural Networks (CNNs) or Transformers have introduced novel technical avenues for style transfer. AdaIN \cite{huang2017arbitrary} uses the VGG network as an image feature encoder to align style and content by adjusting the mean and variance within the feature space. AdaAttN \cite{liu2021adaattn} dynamically adjusts attention based on the shallow and deep features of content and style images, achieving more precise feature distribution matching. CLIPstyler \cite{kwon2022clipstyler} integrates a pre-trained CLIP model with a patch-based text-image matching loss to effectively maintain content structure while introducing style features. The evolution of style transfer techniques is further exemplified by methods like CAST \cite{zhang2022domain}, which improves style transfer accuracy through contrastive learning by analyzing similarities and differences among multiple styles. The success of visual Transformers \cite{dosovitskiy2020image} in the image domain has led to the development of Transformer-based methods like StyTr2 \cite{deng2022stytr2}, demonstrating their efficacy in generating stylized content.

\subsection{Diffusion models for image style transfer}
\label{sec:Diffusion models for image style transfer}

Denoising diffusion probabilistic models (DDPMs) \cite{sohl2015deep,ho2020denoising} model the transition from real data distribution to Gaussian noise distribution through a two-phase process: forward diffusion and reverse sampling. During the forward diffusion phase, DDPM introduces Gaussian noise $\epsilon\sim\mathcal{N}(0,I)$ with a predetermined variance schedule $\beta_t\sim(0,1)$, incrementally blending it into the source clean sample $x_0$ to generate a series of progressively noisy samples $x_t$, as follows:
\begin{equation}
x_t=\sqrt{\bar{\alpha_t}}x_0+\sqrt{1-\bar{\alpha}_t}\epsilon,
\end{equation}

\noindent where $t$ ranges from 1 to $T$, with $\alpha_t=1-\beta_t$ and $\bar{\alpha_t}=\prod_{i=0}^t\alpha_i$. During the reverse phase, the noisy sample $x_T$ is gradually restored to the clean sample $x_0$ by a neural network $\epsilon_\theta$ that predicts and removes noise step-by-step. The Denoising diffusion implicit model (DDIM) \cite{song2021denoising} enhances content preservation during this process with a deterministic sampling formula:
\begin{equation}
x_{t-1} = \sqrt{\alpha_{t-1}}\hat{x}_{0,t} (x_t) + \sqrt{1 - \bar{\alpha}_{t-1}-\sigma^2_t}\epsilon_\theta(x_t, t)+\sigma^2_t\epsilon,
\end{equation}

\noindent setting $\sigma_t=0$ ensures that random Gaussian noise $\epsilon$ does not affect $x_{t-1}$, thus maintaining the content in $x_0$. The denoised image $\hat{x}_{0,t}$ is defined as:
\begin{equation}
\hat{x}_{0,t} (x_t) := \frac{x_t - \sqrt{1 - \bar{\alpha}_t}\epsilon_\theta(x_t, t)}{\sqrt{\alpha_t}}.
\end{equation}

The DDIM deterministic sampling process also preserves the style information in the source sample. Therefore, we set $\sigma_t$ as follows to control the influence of the noise during the reverse process:
\begin{equation}
\sigma_t = \sqrt{\frac{(1 - \bar{\alpha}_{t-1})}{(1 - \bar{\alpha}_t)}}\sqrt{1 - \frac {\bar{\alpha}_t}{\bar{\alpha}_{t-1}}}.
\end{equation}

To introduce specific style information, gradient data from the classifier is incorporated during the reverse sampling process, enabling the stylization according to the following formula \cite{dhariwal2021diffusion}:
\begin{equation}
\hat{x}_{0,t} (x_t) = \hat{x}_{0,t} (x_t) + \nabla_x L_{\text{total}}(x) \rvert_{x=\hat{x}_{0,t} (x_t)},
\end{equation}

\noindent where the $L_{\text{total}}$ includes a style control loss $L_{{sty}}$ and a content control loss $L_{{cont}}$, as follows:
\begin{equation}
  L_{\text{total}}=L_{{sty}}+L_{{cont}}.
  \end{equation}

Diffusion models (DMs) have become prominent in fields such as computer vision and specifically in image style transfer, due to their ability to generate high-quality images. These style transfer methods fall into two categories: image-guided and text-guided. 

Style image-guided methods have been at the forefront of advancing the capabilities of image style transfer, offering powerful stylization techniques. However, the effectiveness of these methods often comes at the cost of requiring either a substantial collection of style images like the WikiArt dataset \cite{artgan2018}, or demanding model training or fine-tuning for each specific style image, which is time-consuming. The process embeds the visual features of the style image into the output domain of the model. For instance, the Dual Diffusion Implicit Bridge (DDIB) \cite{su2022dual} trains two DMs independently in the content and style domains, respectively. It leverages the source model to obtain latent encodings of the source image and decodes them using the target model to generate the stylized image. However, the need for large-scale datasets and extensive training times presents a significant challenge. 

Alternative approaches like StyleDiffusion \cite{wang2023stylediffusion}  have tackled the issue by handling content extraction and style removal with separate DMs, followed by fine-tuning to learn decoupled content and style representations. Similarly, InST \cite{zhang2023inversion} treats style as a learnable text description, using the power of a CLIP model to extract style features and employing a multi-layer cross-attention mechanism to guide the diffusion process. Preserving structural information of content during style transfer is another crucial aspect addressed by OSASIS \cite{cho2024one}, they propose a structure preserving network to incorporate out-of-domain style reference image. 

While style image-guided methods offer significant stylization capabilities, they typically require extensive training or fine-tuning of the model, which can impact the efficiency on style transfer. Thus, StyleID \cite{chung2024style} and Z-star \cite{deng2024z} propose to inject features extracted from the style image into the denoising process of DMs through self-attention.

However, sourcing suitable style images is often time-consuming and labor-intensive. Furthermore, these methods fail when users wish to generate images in entirely new styles for which no corresponding style images exist. As a result, the focus of research has shifted towards more flexible text-guided style transfer methods. A simple way is to utilize the image-toimage ability of the original DMs \cite{huggingface_diffusers_img2img,podell2024sdxl}, while the content information is not well preserved. DiffuseIT \cite{kwon2022diffusion} utilizes intermediate keys extracted by the ViT model along with [CLS] categorical markers, allowing for a nuanced control over the style of generated image and content alignment. Building upon CLIPStyler \cite{kwon2022clipstyler}, ZeCon \cite{yang2023zero} introduces a patch-based contrastive loss that applies during the reverse sampling phase of a pre-trained diffusion model. The technique aims to produce images that are semantically similar to the source, enhancing the content fidelity. However, ZeCon faces challenges with information interaction across the image, sometimes resulting in content missing or inconsistent style application across different regions. Further refining the approach, FreeStyle \cite{he2024freestyle} employs a two-branch encoder diffusion model that distinctly separates and then recombines style guidance and content information during the decoding phase. The method significantly advances style transfer flexibility by allowing distinct handling of style and content. However, FreeStyle encounters difficulties in maintaining the integrity of content from non-diffusion model generated source images.

\subsection{The Shape Space Theory}
\label{sec:The Shape Space Theory}

The Shape Space theory, originally introduced by Kendall in 1984 \cite{kendall_shape_1984}, has been a foundational concept in geometric data analysis. It defines shape as the geometric information that remains when positional, scaling, and rotational effects are eliminated.

In recent years, the Shape Space theory has found applications across diverse domains, leading to significant advancements. For example, Kilian et al. \cite{kilian2007geometric} utilize Geodesic interpolation within the Shape Space to enable the continuous deformation of 3D models. This application highlights the potential of Shape Space for smooth transitions between different shapes. Building on this foundation, Han et al. \cite{han2010recognition,han2014recognizing} develope an object recognition method by projecting object contours into the Pre-Shape Space and identifying Geodesic curves that align with potential shapes of specified object types. The method demonstrates the utility of Shape Space in recognizing and categorizing various shapes based on their geometric properties. Similarly, Paskin et al. \cite{paskin2022kendall} apply the Shape Space theory to biological data by projecting 3D landmarks of shark bones into the Shape Space. The approach allows them to infer the 3D poses of sharks from 2D images using a Geodesic surface. Incorporating Shape Space theory with deep learning, Friji et al. \cite{friji2021geometric} achieve state-of-the-art results in human pose recognition tasks. By combining geometric analysis with neural networks, they enhance the precision and robustness of pose estimation algorithms. Further extending the application of Shape Space, FAGC \cite{han2023fagcfeature} and ITBGS \cite{han2024fewshot} pioneered the projection of image features into the Pre-Shape Space to construct Geodesic curves and surfaces for feature augmentation purposes. The innovative approach enabled the generation of new data points, improving the diversity and quality of training datasets in various machine learning tasks. Despite these advancements, the application of Shape Space theory in the context of diffusion model-based image style transfer remains unexplored.

In a two-dimensional Euclidean space, a shape $P$ can be represented by a set of landmarks, defined as $P=\{p_1(x_1, y_1), ... , p_k(x_k, y_k)\} \in \mathbb{R}^{2\times k}$. However, projecting $P$ into the Shape Space involves operations in complex domains. Therefore, most studies focus on the Pre-Shape Space. The projection of $P$ into the Pre-Shape Space is achieved by a mean-reduction operation $\mathcal{Q}(\cdot)$ and a normalization operation $\mathcal{V}(\cdot)$ to obtain the Pre-Shape $\tau$:
\begin{equation}
P'=\mathcal{Q}(P)=\{p'_i=(x_i-\bar{x}, y_i-\bar{y})\},
\end{equation}

\noindent and
\begin{equation}
\tau=\mathcal{V}(\mathcal{Q}(P))=\mathcal{V}(P') = \frac{P'}{\|P'\|},
 \label{eq:normalization}
\end{equation}

\noindent where $i=1,...,k$ and $k$ denotes the number of landmarks. $||{\cdot}||$ denotes the Euclidean norm.

The Pre-Shape Space can be conceptualized as a hypersphere, where a point on the hypersphere is achieved through the above projection process. Some methods were proposed by Han et al. \cite{han2010recognition,han2014recognizing} to generate new Pre-Shapes from the Geodesic curve built with twoPre-Shapes. The Geodesic curve is derived from the following formula when provided two Pre-Shapes $\tau_1$ and $\tau_2$ in the Pre-Shape Space:
\begin{equation}
\mathbb{G}_{cur}\left(\tau_1,\tau_2\right)\left(s\right)=\left(cos(s)\right)\tau_1+\left(sin(s)\right)\frac{\tau_2-\tau_1cos(d({\tau_1,\tau_2}))}{sin(d({\tau_1,\tau_2}))},
 \label{eq:geo_line}
\end{equation}

\noindent where $d({\tau_1,\tau_2})=\arccos(\tau_1 \cdot \tau_2)$, representing the Geodesic distance between $\tau_1$ and $\tau_2$. $\cdot$ indicates the dot product. The radian $s$, $0\le s\le d({\tau_1,\tau_2})$, controls the Geodesic distance between the newly generated Pre-Shape and $\tau_1$. By gradually increasing $s$, a series of progressively changing Pre-Shapes can be generated.

When dealing with multiple Pre-Shapes, generating new data using using the above formula is limited because it only utilizes two of the total Pre-Shapes, which might not adequately represent the actual data distribution. To address this limitation, various methods have been developed to fully capture the underlying data distribution within the Pre-Shape Space, including seeking an optimal Geodesic curve \cite{han2010recognition,han2023fagcfeature} and approximating Geodesic surfaces using multiple Geodesic curves \cite{han2014recognizing,paskin2022kendall,han2024fewshot}. Constructing a Geodesic surface maximizes the preservation of information from all extracted features. Therefore, we use the method of approximating the Geodesic surface by multiple Geodesic curves to ensure information interaction among all image patches.

\section{Methodology}
\label{sec:Methodology}

\begin{figure*}[htbp]
  \begin{center}
  \includegraphics[width=0.8\linewidth]{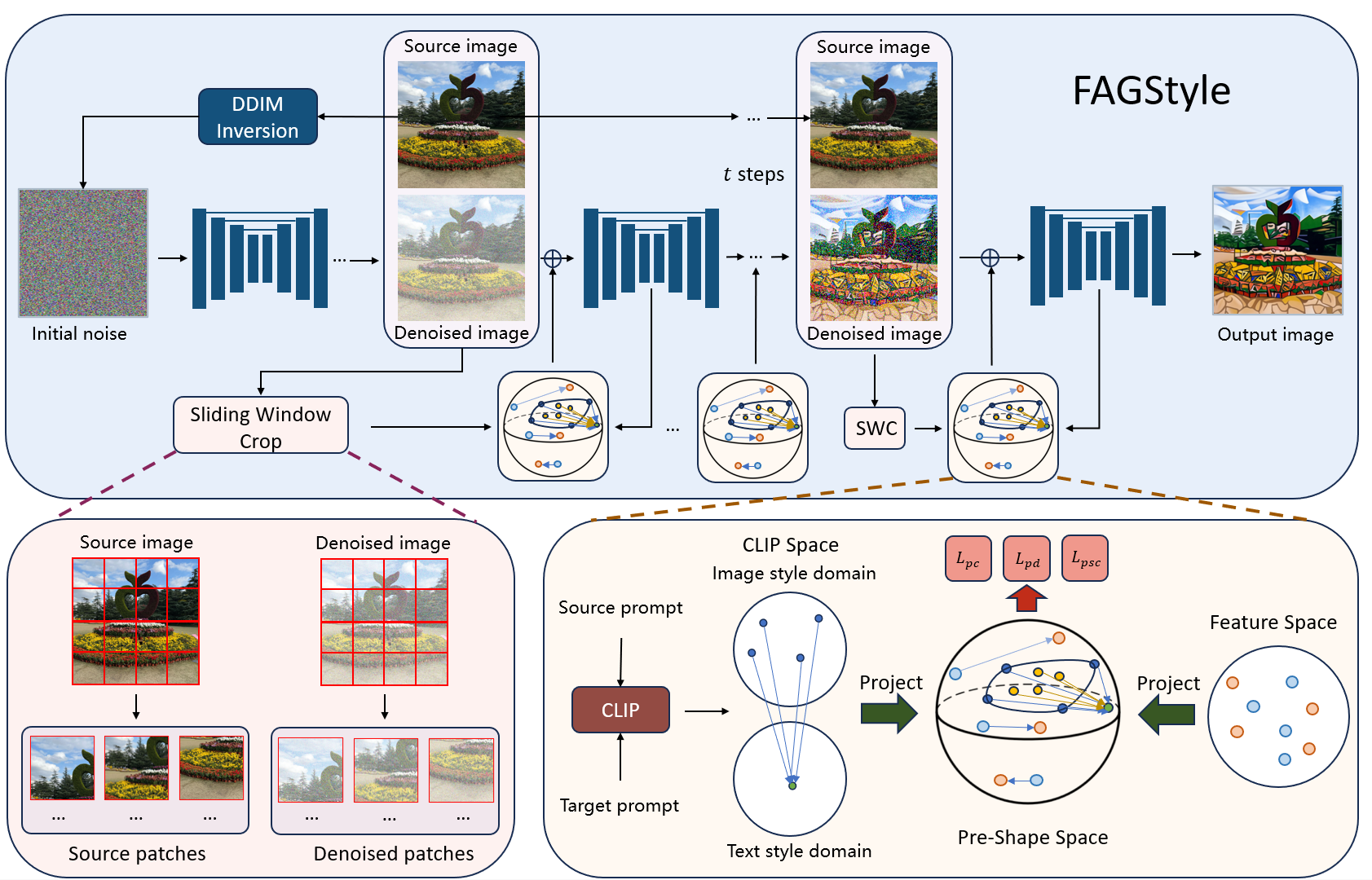}
  \end{center}\caption{The flowchart of our proposed FAGStyle. To better guide the diffusion model for style transfer, we improve the style control loss and content control loss by employing the Sliding Window Crop (SWC) and Feature Augmentation on Geodesic Surface (FAGS) strategies. The gradients of these improved losses is added to the denoised image at each time step during the inference of the diffusion model to guide the style transfer process.} 
  \label{fig:flowchart}
  \end{figure*}

Image style transfer seeks to apply a specific style, conveyed through either image or text, to a source image while preserving its original content. When employing diffusion models for style transfer, a common approach involves initially adding noise to the source image and then incorporating the gradients of style and content control losses into the denoised image during the sampling process as described in Section \ref{sec:Diffusion models for image style transfer}: Diffusion models for image style transfer. We adhere to this paradigm and enhance the existing style and content control losses to achieve a better equilibrium between stylization and content preservation. Given that some newly imagined styles are more readily described using text than by sourcing corresponding style images, we utilize text to represent the style that needs to be transferred.

The flowchart of our proposed FAGStyle is shown in Fig. \ref{fig:flowchart}. In the following subsections, we will introduce our modified style control loss and content control loss in detail.




\subsection{Style Control Loss}
\label{sec:Style Control Loss}

\subsubsection{Sliding Window Crop}
\label{sec:Sliding Window Crop}

Leveraging the powerful multimodal feature extraction capability of CLIP, most existing style control losses are based on features obtained by CLIP extraction. This includes global CLIP loss \cite{Patashnik_2021_ICCV} and directional CLIP loss \cite{gal2022stylegan}. While previous work extracted CLIP features from the entire image for style control loss, CLIPStyler \cite{kwon2022clipstyler} found that using features from a set of randomly cropped patches from the image can achieve better style transfer results. They concluded that patch-based CLIP loss provides spatially invariant information compared to global CLIP loss.

However, randomly cropped patches may concentrate in certain regions, leading to over-stylization of specific areas and under-stylization of others. Also, using random crop may compromise the main structure of the source content, as shown in Row 1 of Fig. \ref{fig:ablation_quali} in the ablation study. To address the above issues, we introduce a sliding window crop (SWC) approach to crop overlapping patches, ensuring that the union of all patches covers the entire image. The approach allows the style control loss to consider all regions of the image while still benefiting from the spatially invariant information provided by using cropped patches to calculate style control loss.

For the denoised image $\hat{x}_{0,t} (x_t)$ and the source image $x_0$ at step $t$, we assume that they are both resized to $H\times W$ with $H=W$. We set the number of patches to $n$ as a hyperparameter, ensuring that the images are uniformly cropped into $n_w=n_h=\sqrt{n}$ square patches, each with height and width $H_p=W_p=W/(n_w+1)=H/(n_h+1)$. The stride of the sliding window is set to $s=H_p/2=W_p/2$, so that each patch has certain overlapping portions with adjacent patches, ensuring information interaction among them. The coordinates $(e,f)$ of the upper left corner of each patch $\hat{x}^i_{0,t}$ and $x^i_0$ obtained by SWC for image $\hat{x}_{0,t}$ and $x_0$, respectively, are given by:
\begin{equation}
(e,f)=(\left\lfloor \frac{i}{n_h} \right\rfloor \cdot s, (i\mod n_w)\cdot s),
\end{equation}

\noindent where $i=0,1,... ,n-1$, $\lfloor\cdot\rfloor$ denotes the floor function, and $\text{mod}$ denotes modulo operation. The proposed SWC approach dynamically adjusts the window size and stride according to the input image size and the number of required patches. This ensures the overlapping portions among adjacent patches for effective information interaction, resulting in more uniform and consistent style transfer across the entire image.

\subsubsection{Feature Augmentation on Geodesic Surface}
\label{sec:Feature Augmentation on Geodesic Surface}

While the SWC ensures some degree of information interaction among adjacent patches, it does not address the lack of mutual information among non-adjacent patches. This is particularly problematic when a large number of patches are used, resulting in significant spatial distances between the first and last patch. The lack of interaction can lead to over-stylization or under-stylization in different parts of the image, affecting the overall quality of the generated images.

To address this issue, we adopt the Feature Augmentation on Geodesic Surface (FAGS) strategy \cite{han2024fewshot}. Following FAGS, the features of all patches obtained by SWC are projected into the Pre-Shape Space to build a Geodesic surface. Any point on this Geodesic surface represents a augmented new feature. The augmented feature contains weighted information from all other cropped patches. We then calculate the style control loss based on these augmented feature points on the built Geodesic surface.

We use the image encoder $E_{\text{img}}$ of the pre-trained CLIP model to extract features $\bar{x}_t$ and $\bar{x}_0$ from the denoised image patches and the source image patches, respectively:
\begin{equation}
\bar{x}_t = \left\{ E_{\text{img}}(\hat{x}^i_{0,t}) \mid E_{\text{img}}(\hat{x}^i_{0,t}) \in \mathbb{R}^{c \times h \times w}, i = 1, 2, \dots, n \right\},
\end{equation}

\noindent and
\begin{equation}
\bar{x}_0 = \left\{ E_{\text{img}}(x^i_{0}) \mid E_{\text{img}}(x^i_{0}) \in \mathbb{R}^{c \times h \times w}, i = 1, 2, \dots, n \right\}.
\end{equation}

Since the Geodesic surface is defined in the Pre-Shape Space, the extracted features must be projected into the Pre-Shape Space. The operation $\mathcal{R}(\cdot)$ resizes the feature dimensions from $c \times h \times w$ to $2 \times (chw/2)$, representing the set of $chw/2$ landmarks in 2D space. This allows the features to be viewed as a shape in 2D space for subsequent projection into the Pre-Shape Space. Using the mean-reduction and normalization operations from Formula 5 and Formula 6, the projection into the Pre-Shape Space is defined as $f_p(\cdot)=\mathcal{V}(\mathcal{Q}(\mathcal{R}(\cdot)))$.

A Geodesic surface $\mathbb{G}_{surf}$ on the Pre-Shape Space can be defined as follows \cite{paskin2022kendall,han2024fewshot}:
\begin{equation}
\mathbb{G}_{surf}(\tau,\omega)=\mu_n,
\end{equation}

\noindent and
\begin{equation}
\mu_j=\mathbb{G}_{cur}(\mu_{j-1}, \tau_j)(\frac{\omega_j}{\sum^j_{i=1}\omega_i}),\ {\rm where}\ j=2,...,n,\
\end{equation}

\noindent where $\mu$ is a vector and also a Pre-Shape on the Geodesic surface $\mathbb{G}_{surf}$. $\tau\triangleq {[\tau_1, ... ,\tau_n]^T}$ and $\omega\triangleq{[\omega_1,... ,\omega_n]^T}$ represent the given set of vectors and weights, respectively. $n$ represents the number of given vectors, which is equivalent to the number of patches in this case. $\mu_1=\tau_1$, so when $j=n$, a Geodesic surface $\mathbb{G}_{surf}$ can be built using a set of vectors $\tau$ and a set of weights $\omega$. The weights $\omega$ control the information interaction among all patch features.

We construct two Geodesic surfaces $\mathbb{G}_{surf}(f_p(\bar{x}_t),\omega)$ and $\mathbb{G}_{surf}(f_p(\bar{x}_t),\omega)$ on the Pre-Shape Space, using the feature sets from the denoised image patches and the original patches, respectively, as the given vector sets. Multiple augmented feature vectors $\tilde{x}_t^{i}$ and $\tilde{x}_0^{i}$ can be obtained on the corresponding built Geodesic surface by providing $m$ different sets of weights, where $i=1,2, ... ,m$. Here, we set $m=n$, so that the number of augmented features equals the number of cropped patches. Each newly augmented feature vector contains mutual information from all patches, enabling information interaction among non-adjacent patches.

\subsubsection{Improved Style Control Loss}
\label{sec:Improved Style Control Loss}

The improved style control loss function of our method combines the SWC approach with the FAGS strategy on the Pre-Shape Space, and can be defined as follows:
\begin{equation}
L_{sty}=\lambda_{pc}L_{pc}(\hat{x}_{0,t}, p_{tgt})+\lambda_{pd}L_{pd}(\hat{x}_{0,t}, x_0, p_{tgt}, p_{src}),
\end{equation}

where $L_{pc}$ and $L_{pd}$ integrate FAGS strategy based on the SWC, improving on the existing PatchCLIP Loss \cite{kwon2022clipstyler} and Patch Directional Loss \cite{yang2023zero}. Here, $p_{tgt}$ and $p_{src}$ are the text prompts for style reference and the text description of the source image style, respectively. $p_{src}$ is set to "Photo" when a real photograph is used as the input image. $\lambda_{pc}$ and $\lambda_{pd}$ are hyperparameters corresponding to the weights of $L_{pc}$ and $L_{pd}$, respectively.

The $L_{pc}$ calculates the Geodesic distance between the target CLIP feature of $p_{tgt}$ and the augmented source patch feature $\tilde{x}_t^{i}$ obtained from the built Geodesic surface, as shown in the following formula:
\begin{equation}
L_{pc}(\hat{x}_{0,t}, p_{tgt})= \frac{1}{n}\sum \limits^n_{i=1} d(\tilde{x}_t^{i}, f_p(E_{\text{txt}}(p_{tgt}))),
\end{equation}

\noindent where $E_{\text{txt}}$ denotes the text encoder of CLIP. Since $\tilde{x}_t^{i}$ is located in the Pre-Shape Space, it is also necessary to project the target CLIP features $E_{\text{txt}}(p_{tgt})$ into the Pre-Shape Space for the calculation of the Geodesic distance. 

The $L_{pd}$ aligns the Pre-Shape Space direction between source image/text style features and target image/text style features, as shown in the following formula:
\begin{equation}
L_{pd}(\hat{x}_{0,t}, x_0, p_{tgt}, p_{src})=\frac{1}{n}\sum \limits^n_{i=1}(1-\frac{\bigtriangleup I_i\cdot\bigtriangleup T}{|\bigtriangleup I_i|\cdot|\bigtriangleup T|}),
\end{equation}

\noindent where we utilize the augmented source patch feature $\tilde{x}_0^{i}$ and augmented target patch feature $\tilde{x}_0^{i}$ from the built Geodesic surface to demonstrate their Pre-Shape Space direction:
\begin{equation}
\bigtriangleup I_i=\tilde{x}_t^{i}-\tilde{x}_0^{i}.
\end{equation}

Similarly, we project the source text feature $E_{\text{txt}}(p_{src}))$ and the target text feature $E_{\text{txt}}(p_{tgt}))$ into the Pre-Shape Space to calculate the Pre-Shape Space direction between them:
\begin{equation}
\bigtriangleup T=f_p(E_{\text{txt}}(p_{tgt})))-f_p(E_{\text{txt}}(p_{src}))).
\end{equation}

Our improved style control loss ensures information interaction among adjacent and long-distance patches while providing spatially invariant information, resulting in uniform and consistent style transfer across the entire image.

\subsection{Content Control Loss}
\label{sec:Content Control Loss}

Ensuring both content preservation and effective stylization during the style transfer process is challenging. While the previous section improve the style control loss to enhance uniformity and consistency of style, a new content control loss is introduced here to further balance stylization and content preservation.

\subsubsection{Pre-Shape Self-correlation Consistency Loss}
\label{sec:Pre-Shape Self-correlation Consistency Loss}

Although the improved style control loss enhances style expression, it may sometimes weaken content preservation. The self-correlation consistency loss proposed in RSSA \cite{xiao2022few} ensures the intrinsic structural consistency between the target and source images by analyzing image features and calculating self-correlation matrices. ITBGS \cite{han2024fewshot} further enhances this by correlating the consistency between target image features from the built Geodesic surface and source image features in the Pre-Shape Space, proposing the Geodesic self-correlation consistency loss. To improve the ability of content preservation during stylization, we introduce the Pre-Shape self-correlation consistency loss $L_{psc}$.

Recent work by Baranchuk et al. \cite{baranchuk2021label} has demonstrated that the intermediate features obtained by the U-Net noise predictor $\epsilon_\theta$ in the diffusion model contain spatial information of content in image. Since the self-correlation consistency loss is calculated at the feature level, we employ $\epsilon_\theta$ as the feature extractor. At each step of the reverse denoising process, both the source image $x_0$ and the denoised image $\hat{x}_{0,t}$ are input into $\epsilon_\theta$. The encoder part $E_{\epsilon_\theta}$ of $\epsilon_\theta$ can be used to extract the feature maps of $x_0$ and $\hat{x}_{0,t}$ respectively. 

In the ITBGS setup, multiple source and target images are provided. They utilize the augmented source and target image features through FAGS to calculate the Geodesic self-correlation consistency loss. However, our setup involves only one source image $x_0$ and one target image $\hat{x}_{0,t}$. Since cropping image into patches may disrupt content integrity, we do not apply FAGS to our content control loss. Instead, we directly project the features of the source and target images into the Pre-Shape Space as follows:
\begin{equation}
z^l_0=f_p(E_{\epsilon_\theta}(x_0))\in \mathbb{R}^{c \times h \times w},
\end{equation}

\noindent and
\begin{equation}
\hat{z}^l_{0,t}=f_p(E_{\epsilon_\theta}(\hat{x}_{0,t}))\in \mathbb{R}^{c \times h \times w}.
\end{equation}

After obtaining the source image features $z^l_0$ and target image features $\hat{z}^l_{0,t}$ in the Pre-Shape Space, we calculate their self-correlation matrices. $z^l_0(u,v)$ and $\hat{z}^l_{0,t}(u,v)$ indicate vectors in $c$ dimensions at the position $(u,v)$ of $z^l_0$ and $\hat{z}^l_{0,t}$, respectively. The cosine similarity $C^{z_0^l}_{u,v}(a,b)$ between vector at position $(u,v)$ of $z^l_0$ and the corresponding vector at position $(a,b)$ is calculated as follows:
\begin{equation}
C^{z_0^l}_{u,v}(a,b)=\frac{<z_0^l(u,v), z_0^l(a,b)>}{\|z_0^l(u,v)\|\cdot\|z_0^l(a,b)\|},
\end{equation}

\noindent traversing all corresponding positions in the feature space yields the self-correlation matrix $C^{z_0^l}_{u,v}$ for $z^l_0(u,v)$. Similarly, we obtain the self-correlation matrix $C^{\hat{z}^l_{0,t}}_{u,v}$ for $\hat{z}^l_{0,t}(u,v)$. The Pre-Shape self-correlation consistency loss $L_{psc}$ is then calculated as follows:
\begin{equation}
L_{psc}(x_0, \hat{x}_{0,t})=\mathbb{E}_{x_0}\sum_l\sum_{u,v}L_{s\ell1}(C^{z^l_0(u,v)}_{u,v}, C^{\hat{z}^l_{0,t}}_{u,v}),
\end{equation}

\noindent where $l$ iterates through selected layers of the noise predictor $\epsilon_\theta$, $(u,v)$ traverses through all positions in the feature space, and $L_{s\ell1}(\cdot)$ denotes the smooth-$\ell$1 loss function \cite{ren2015faster}.

\subsubsection{Improved Content Control Loss}
\label{sec:Improved Content Control Loss}

By introducing $L_{psc}$ to maintain the structural consistency in the target image, the content of the source image is better preserved. The content control loss $L_{\text{cont}}$ with the addition of $L_{psc}$ is constructed as follows:

\begin{IEEEeqnarray}{rCl}
  L_{\text{cont}} &=& \lambda_{ps}L_{psc}(x_0, \hat{x}_{0,t}) + \lambda_{z}L_{ZeCon}(x_0, \hat{x}_{0,t}) \nonumber \\
  &+& \lambda_{v}L_{VGG}(x_0, \hat{x}_{0,t}) + \lambda_{m}L_{MSE}(x_0, \hat{x}_{0,t}).
  \end{IEEEeqnarray}

We additionally introduce $L_{ZeCon}$, which computes the cross-entropy loss between the features extracted from $x_0$ and $\hat{x}_{0,t}$ \cite{yang2023zero}. $L_{VGG}$ minimizes the mean square error the VGG feature maps of $x_0$ and $\hat{x}_{0,t}$. $L_{MSE}$ calculates the $\ell2$ loss at the pixel level of the two images. The weights $\lambda$ of each loss function are also fixed hyperparameters.

\section{Experiments}
\label{sec:Experiments}

\subsection{Implementation Details}
\label{sec:Implementation Details}

All experiments in this section are conducted with fixed hyperparameters, as described in Section \ref{sec:Hyperparameters setting}: Hyperparameters Setting. We utilize the unconditional diffusion model \cite{dhariwal2021diffusion} pre-trained on ImageNet with $\rm 256\times256$ image size as the generative model. The pre-trained CLIP \cite{radford2021learning} serves as the text encoder. For the forward diffusion process, we apply DDIM inversion, setting $t_0=25$ to obtain the noised $x_{t_0}$ while respacing the default time step from $T=1000$ to $T'=50$ \cite{yang2023zero}. During the reverse sampling process, we adopt the DDPM strategy for better stylization results. 

\subsection{Experimental Results}
\label{sec:Experimental Results}

To verify the robustness and generality of our proposed method, we compare it with eight state-of-the-art (SOTA) methods using 25 source content images and 55 different styles. The source content images are collected from ImageNet \cite{jia2009imagenet} and web collections. These images span various domains, including landscapes, animals, transportation, food, and architecture. The styles include 30 imagined styles, which are challenging to find suitable reference images for, and 25 common styles. 

\begin{figure*}[htbp]
  \begin{center}
  \includegraphics[width=0.8\linewidth]{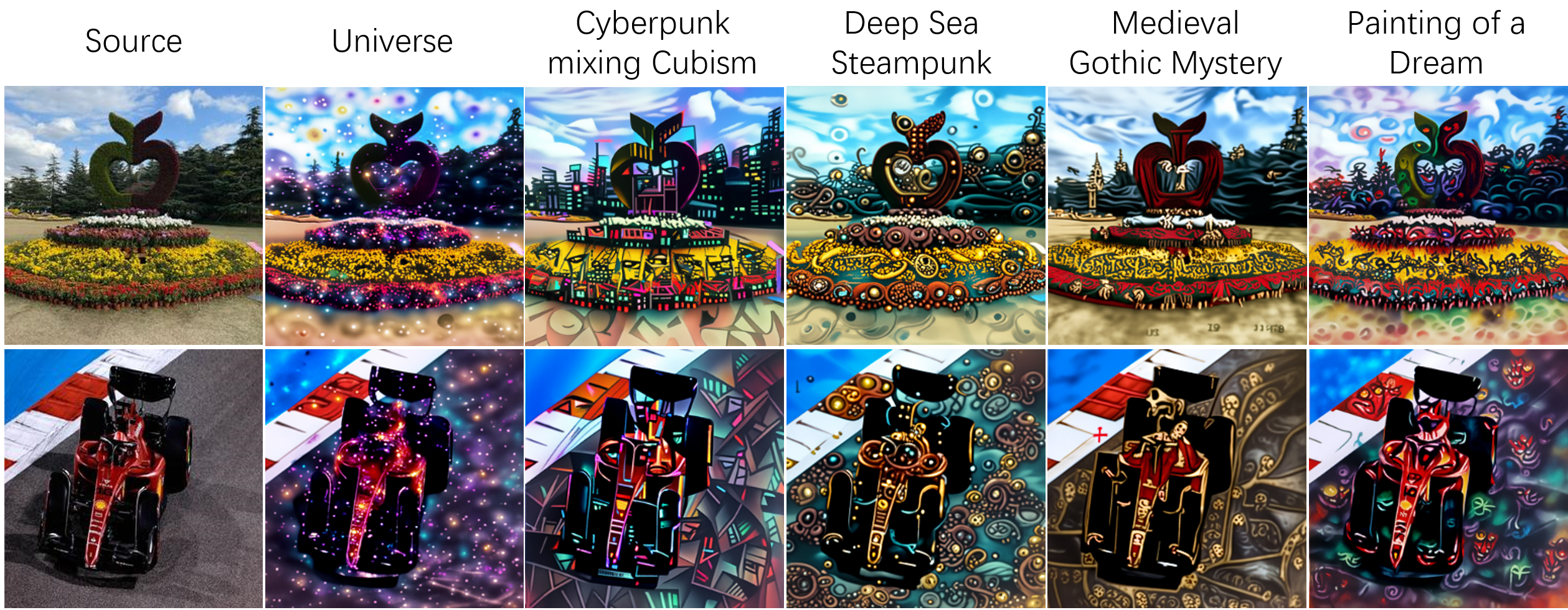}
  \end{center}\caption{Text-guided Image Style transfer results using FAGStyle.} 
  \label{fig:first_image}
  \end{figure*}

Fig. \ref{fig:first_image} shows that our method achieves outstanding results in both imagined and common styles. In addition, we perform comparison experiments with the eight SOTA methods in terms of both qualitative and quantitative evaluations. The comparison methods include five text-guided style transfer methods: SDXL \cite{podell2024sdxl}, CLIPstyler \cite{kwon2022clipstyler}, DiffuseIT \cite{kwon2022diffusion}, ZeCon \cite{yang2023zero}, and Freestyle \cite{he2024freestyle}. Additionally, three image-guided style transfer methods are included: Stytr2 \cite{deng2022stytr2}, CAST \cite{zhang2022domain}, and InST \cite{zhang2023inversion}. We employ publicly available implementations of all methods, using their recommended configurations.

\subsubsection{Comparison of Methods in Imagined Styles}
\label{sec:Comparison of Methods in Imagined Styles}

\begin{figure*}[htbp]
  \begin{center}
  \includegraphics[width=1\linewidth]{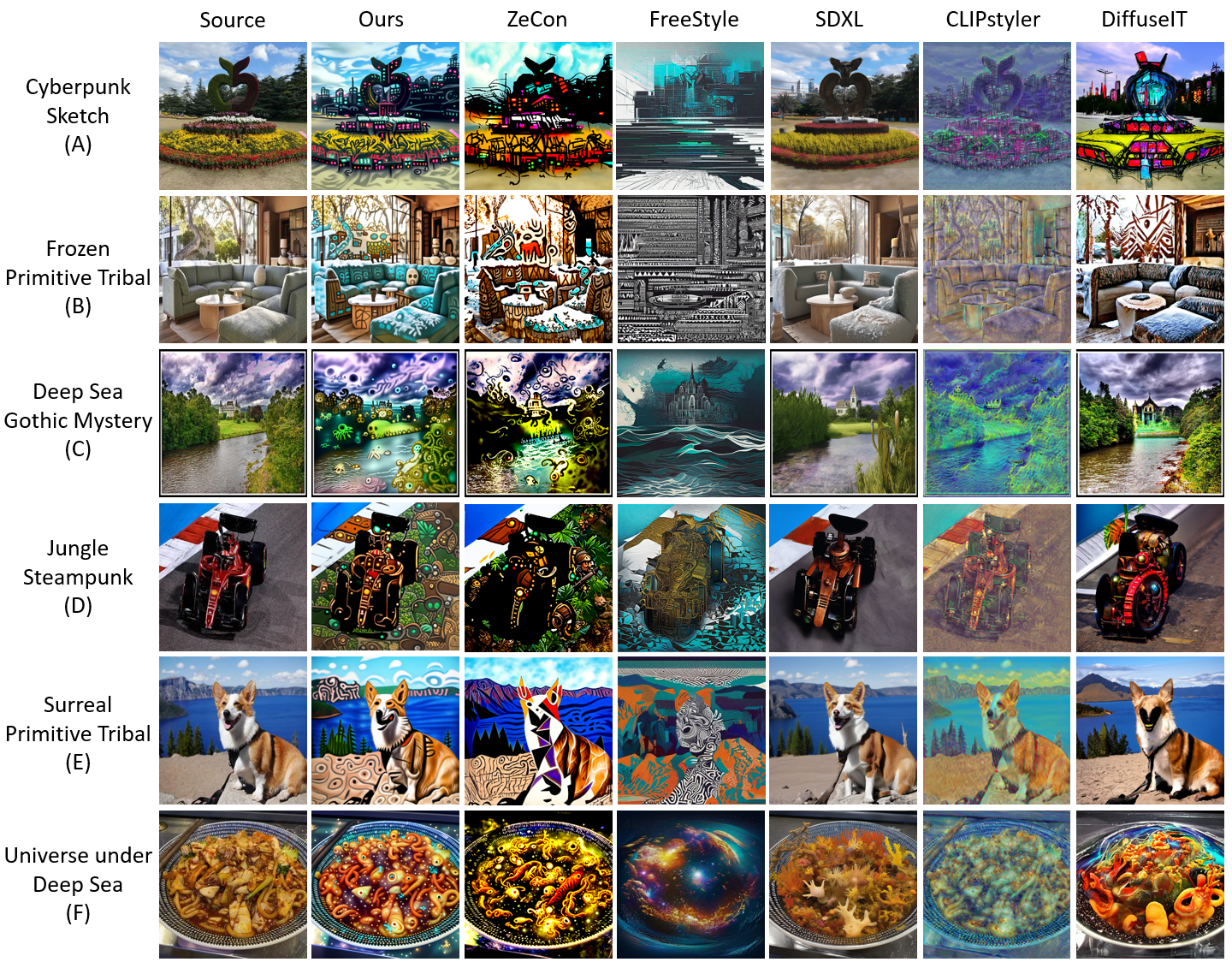}
  \end{center}\caption{Qualitative comparison of our proposed method with other text-guided style transfer methods in imagined styles.} 
  \label{fig:quali_comparison_imagined}
  \end{figure*}

Our method employs a text-guided approach, enabling users to articulate novel, imagined styles through textual descriptions and transfer these onto a source content image. Such styles, like "Painting of chaos," often lack corresponding style images, making them challenging to represent visually. By merging two common styles, we have devised a total of 32 imagined styles, including "Deep Sea Gothic Mystery" and "Jungle Steampunk." For these imagined styles, we exclusively utilize text-guided style transfer methods for our comparisons, as illustrated in Fig. \ref{fig:quali_comparison_imagined}. 

DiffuseIT, originally designed for image editing, when adapted to style transfer tasks, tends to preserve the real photographic styles of the input images while merely altering the content. For instance, in Row D, the race car in the source image is adapted to a steampunk style with an additional tree to align with the Jungle theme, yet no overall stylistic alterations are made, leading to under-stylization. CLIPstyler modifies the overall color tone and texture of the image during the stylization process, akin to applying a filter. While this method achieves a more consistent style, the extent of content stylization remains modest. For SDXL img-to-img, we adjusted the strength and classifier-free guidance scale to 0.6 and 7.5, respectively. Occasionally, SDXL significantly alters the content (Rows A, C, D, and F), but in Rows B and E, the images remain largely unchanged from the source. This variability indicates that SDXL is sensitive to parameter settings and often requires multiple adjustments to tune the parameters appropriately for different content and style prompts. Similarly to DiffuseIT, SDXL only modifies parts of the content, facing challenges with achieving full stylization of the entire image. FreeStyle, in some instances, almost completely loses the content information specified by certain style descriptions (Rows A and C), and in other scenarios, it retains merely the outline shape of the content without any semantic details (Rows B, D, E, and F). ZeCon, on the other hand, not only stylizes the content within the image but also preserves aspects like the color tone of the source image. However, it occasionally omits some details of the source content, such as the house details outside the window in Row B and the front suspension of the race car in Row D.

In comparison, our method effectively transforms content while preserving the source contours, enabling stylistic edits. For instance, in Row F, FAGStyle converts rice noodles into deep-sea creatures adorned with stars. Textures are also modified, as shown in Row B, FAGStyle alters a racing car photo to exhibit a steampunk texture. Furthermore, our method ensures uniform stylization across the entire image. For example, in Row C, it seamlessly blends the sky, forest, and water into a Deep Sea mixed with Gothic style, while preserving the architectural details clearly.

\begin{table}[htbp]
  \centering
    \caption{Quantitative comparisons in imagined styles. FS represents FreeStyle, CS denotes ClipStyler, and DIT stands for DiffuseIT. The best results are highlighted in bold while the second-best results are underlined.} 
  \begin{tabularx}{\linewidth}{>{\centering\arraybackslash}m{1.4cm}|>{\centering\arraybackslash}X>{\centering\arraybackslash}X>{\centering\arraybackslash}X>{\centering\arraybackslash}X>{\centering\arraybackslash}X>{\centering\arraybackslash}X}
  \toprule
  Metrics & Ours            & ZeCon        & FS & SDXL             & CS & DIT \\ \midrule
  PSNR $\uparrow$    & \uline{28.27}   & 27.94    & 27.95   & \textbf{28.82} & 27.89    & 28.21   \\
  SSIM $\uparrow$    & \textbf{0.498} & 0.277       & 0.138    & \uline{0.440}     & 0.297      & 0.202    \\ \midrule
  CLIP-I $\uparrow$  & \textbf{0.334} & \uline{0.317} & 0.268    & 0.224          & 0.273     & 0.252    \\
  CLIP-P $\uparrow$  & \textbf{0.242} & \uline{0.239} & 0.218    & 0.211          & 0.226     & 0.216    \\ \bottomrule
  \end{tabularx}
  \label{tab:quantitative_comparison_imagined}
  \end{table}

To better evaluate the performance of our proposed method, we utilize multiple quantitative metrics, as shown in Table \ref{tab:quantitative_comparison_imagined}. We measured the Peak Signal-to-Noise Ratio (PSNR) and Structural Similarity Index (SSIM) to assess the content similarity between the generated images and the source content images \cite{5596999}. Additionally, we calculated CLIP scores to determine the style similarity between the generated images and style reference prompts. The CLIP-I score, calculated across the entire image, reflects the overall degree of style transfer, while the CLIP-P score, determined by averaging scores across multiple 64 $\times$ 64 patches, assesses the consistency of the style throughout the image. Our method achieved the highest scores in the SSIM, CLIP-I, and CLIP-P metrics, indicating superior performance in content preservation and style transfer consistency. SDXL is good at generating high quality images. The PSNR score of our method is slightly lower than SDXL, but still ranks second. Our method shows only slight improvements over ZeCon in the two CLIP score metrics, suggesting that both methods are proficient in stylization. However, ZeCon falls short in content preservation, as indicated by its lower PSNR and SSIM scores compared to ours. The results demonstrate that our method can effectively balance the stylization and content preservation of the generated images.

\subsubsection{Comparison of Methods in Common Styles}
\label{sec:Comparison of Methods in Common Styles}

\begin{figure*}[htbp]
  \begin{center}
  \includegraphics[width=1\linewidth]{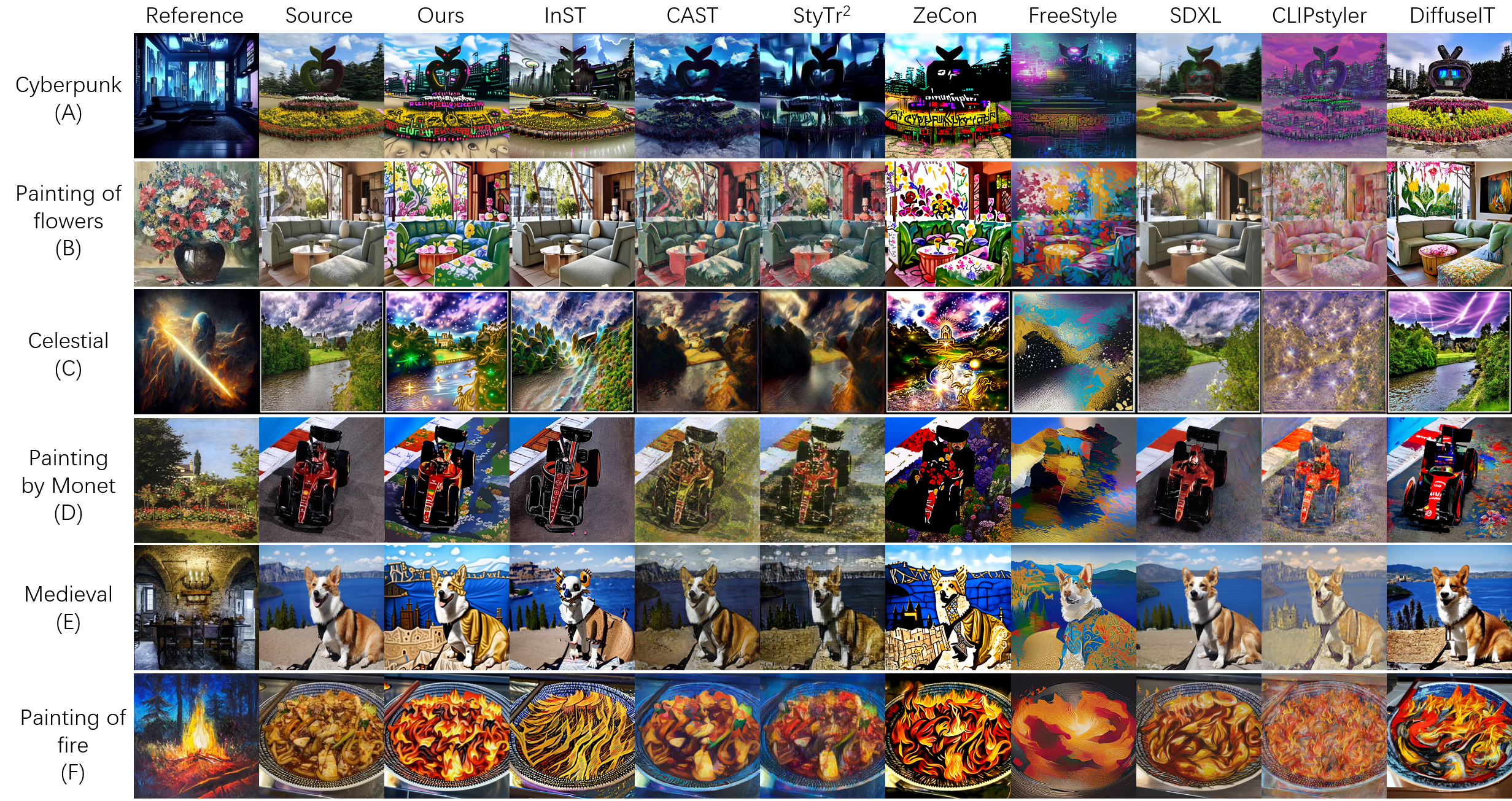}
  \end{center}\caption{Qualitative comparison of our proposed method with other style transfer methods in some of the common styles.} 
  \label{fig:quali_comparison_common}
  \end{figure*}

  

  In this section, we explores a total of 23 common styles, including Cyberpunk, Medieval, etc. Due to the accessibility of appropriate style reference images for the common styles, we not only compare the above 5 text-guided methods but also 3 image-guided techniques. Fig. \ref{fig:quali_comparison_common} demonstrates the qualitative results of our methods against the 8 style transfer methods.

  When dealing with common styles, text-guided methods face challenges similar to those encountered with imagined styles. For instance, DiffuseIT and SDXL sometimes excessively alter the content, as seen in Row A, and in other instances, changes are barely noticeable, such as the minimal transformation of a Corgi in Row E and the conversion of desktop objects into a vase or changing the outside view into flowers in Row B. ClipStyler, altering the texture and tone of the entire image, results in a generally gray and blurry visual experience, which significantly deviates from the original tone and style. Conversely, FreeStyle retains only the contours of the content and fills in the style within these boundaries. ZeCon preserves the semantic information of the content in the stylized images and builds upon it with stylistic modifications, such as the clearly recognizable couch and table in Row B, albeit filled with floral patterns. However, this approach may lose some content outlines, as evidenced by the plant sculpture in Row A and the altered overall structure of the house in Row C.

  For image-guided methods, we utilize style reference images that closely match the text prompts, directing the style transfer. StyTr2 and CAST extract texture and color information from the reference images to perform style transfer. However, using a single image as a guide often fails to fully reflect the style information contained in the text. For example, in Row A, both StyTr2 and CAST merely adjust the color scheme of the source image to match the blue and black tones of the reference without capturing and transferring the architectural style or neon colors characteristic of Cyberpunk. Thus, these image-guided methods are more successful with styles that are closely related to color or texture, such as those shown in Rows B, D, and F. The InST method initially converts the input reference image into a text embedding through textual inversion and utilizes Stable Diffusion as the backbone to achieve style transfer through text-to-image generation. This addition of textual information allows it to transform the forest in the source image into the architectural style of Cyberpunk, as seen in Row A. However, it too can exhibit inconsistent levels of stylization, as shown in Rows D and E, where only the main subjects, the racing car, and the Corgi, are stylized, leaving other parts unchanged.


In contrast, FAGStyle effectively preserves and stylizes each part of the source image content while ensuring the uniform consistency of the overall image style. For example, in Row A, the forest is transformed into Cyberpunk-style architecture. Our method also maintains the content contours while modifying the textures, such as adding vibrant patterns to the sofa in Row B. Furthermore, our method ensures a consistent level of stylization across the entire image, as seen in Row E, not only the Corgi but also the cliffside forest, the ground, and the distant mountains are transformed into the Medieval style.
  


\begin{table*}[htbp]
  \centering
  \caption{Quantitative comparisons in common styles. FS represents FreeStyle, CS denotes ClipStyler, and DIT stands for DiffuseIT. The best results are highlighted in bold while the second-best results are underlined.} 
  \resizebox{\linewidth}{!}{
    \begin{tabularx}{\linewidth}{>{\centering\arraybackslash}X|>{\centering\arraybackslash}X>{\centering\arraybackslash}X>{\centering\arraybackslash}X>{\centering\arraybackslash}X>{\centering\arraybackslash}X>{\centering\arraybackslash}X>{\centering\arraybackslash}X>{\centering\arraybackslash}X>{\centering\arraybackslash}X}
    \toprule
    Metrics & Ours           & InST             & CAST    & $\rm Stytr^2$ & ZeCon           & FS & SDXL          & CS & DIT \\ \midrule
    PSNR $\uparrow$    & 28.29        & \textbf{28.52} & 27.98 & 27.98      & 28.07         & 27.96   & \uline{28.45} & 27.89    & 28.18    \\
    SSIM $\uparrow$    & \textbf{0.51}  & 0.389           & 0.402  & \uline{0.415}  & 0.356          & 0.136    & 0.369        & 0.295     & 0.182    \\ \midrule
    CLIP-I $\uparrow$  & \textbf{0.293}   & 0.229           & 0.242  & 0.237        & \uline{0.292} & 0.260    & 0.252       & 0.266     & 0.252    \\
    CLIP-P $\uparrow$  & \textbf{0.236} & 0.212           & 0.229  & 0.223        & \uline{0.234}    & 0.219     & 0.222        & 0.233     & 0.219    \\ \bottomrule
    \end{tabularx}
  }
  \label{tab:quantitative_comparison_common}
  \end{table*}

We also conduct quantitative experiments with three additional image-guided methods in common styles, as detailed in Table \ref{tab:quantitative_comparison_common}. Similar to results in imagined style, our method outperforms others by achieving the highest scores in SSIM and both CLIP metrics, indicating it excels in maintaining original content and applying consistent styles. In PSNR, we rank third, just behind InST and SDXL. Both of these methods use the Stable Diffusion as backbone, highlighting its capability in producing high-quality images. Despite their strengths, these methods did not match our performance in style transfer metrics. Our method led the CLIP scores, with ZeCon closely behind, which confirms the effectiveness of both in achieving coherent style transfers. However, ZeCon was less effective in preserving content similarity compared to our method. The results demonstrate superior ability of our method to deliver high-fidelity and stylistically consistent image transformations across various styles.


\subsection{Ablation Study}
\label{sec:Ablation Study}

To verify the effectiveness of the proposed improvements to our FAGStyle method regarding style control loss and content control loss, we conduct both qualitative and quantitative ablation study on an image of an apple-shaped plant sculpture as our source content with 5 different styles. We introduce the Sliding Window Crop (SWC) technique and the integration of Feature Augmentation on Geodesic Surface (FAGS) into our style control loss to bolster information interaction among patches. Furthermore, for content control loss, we detail the projection of image features into the Pre-Shape Space to compute self-correlation consistency loss.


\begin{figure*}[htbp]
  \begin{center}
  \includegraphics[width=0.9\linewidth]{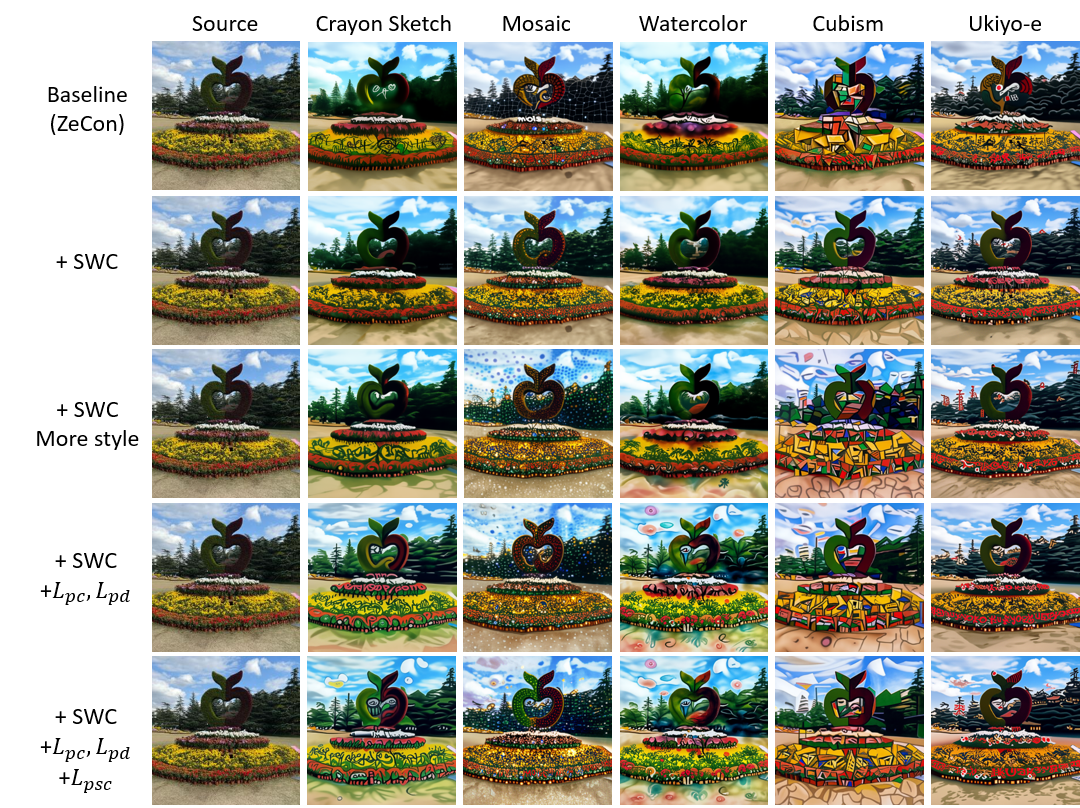}
  \end{center}\caption{The Qualitative ablation study of the proposed improvements in FAGStyle regarding style control loss and content control loss.} 
  \label{fig:ablation_quali}
  \end{figure*}

\begin{table}[htbp]
  \centering
  \caption{The Quantitative ablation study of the proposed improvements in FAGStyle regarding style control loss and content control loss. The best results are highlighted in bold while the second-best results are underlined.} 
  \begin{tabular}{@{}ccc|cc@{}}
  \toprule
  Methods   & PSNR $\uparrow$            & SSIM $\uparrow$           & CLIP-I $\uparrow$         & CLIP-P $\uparrow$          \\ \midrule
  Baseline  & 28.112          & 0.469          & \textbf{0.253} & 0.221           \\
  +SWC      & \textbf{28.274} & \textbf{0.537} & 0.203          & 0.211           \\
  +SWC w/. More Style   & 28.201          & 0.473          & 0.244          & 0.221           \\
  +SWC, $L_{pc}$, $L_{pd}$ & 28.223          & 0.495          & \uline{0.253}    & \textbf{0.223} \\
  +SWC, $L_{pc}$, $L_{pd}$, $L_{psc}$  & \uline{28.238}    & \uline{0.498}    & 0.252          & \uline{0.223}    \\ \bottomrule
  \end{tabular}
  \label{tab:ablation_quanti}
  \end{table}

\subsubsection{Effect of the SWC}
\label{sec:Effect of the SWC}

We transitioned from a random crop to the Sliding Window Crop (SWC) approach to assess its impact on stylization. The baseline method, ZeCon, which employs random cropping, frequently results in the loss of the shape information of the sculpture across various styles such as Crayon Sketch, Mosaic, Watercolor, and Ukiyo-e. Additionally, ZeCon exhibits inconsistencies in overall style transformation. For instance, in the Cubism stylization results, the sculpture is stylized while the woods in the background remain unchanged. Switching to SWC more effectively preserves content, though it initially leads to slight under-stylization, as evidenced in Row 2 of Fig. \ref{fig:ablation_quali} and Table \ref{tab:ablation_quanti}. By increasing the weight of the style control loss, Row 3 of Fig. \ref{fig:ablation_quali} demonstrates enhanced preservation of the source image content along with a more style application compared to baseline. This can also be reflected in the increased PSNR and SSIM, as well as the same CLIP-P shown in Table \ref{tab:ablation_quanti}. With the implementation of SWC, both the woods and the sculpture maintain stylistic coherence, as shown in the Cubism results.



\subsubsection{Effect of the FAGS strategy}
\label{sec:Effect of the FAGS strategy}

As observed in Row 3 of Fig. \ref{fig:ablation_quali}, applying higher weights still sometimes leads to inconsistencies in the stylization across different parts of the image, such as the left and right sections of the woods in the Watercolor stylization. Additionally, increasing the weights tends to compromise content preservation, evidenced by the loss of color information in the sculpture in both the Mosaic and Watercolor stylized images.

This suggests that the original style control loss, even with the implementation of SWC, does not facilitate adequate interaction between patches, resulting in uneven stylization and partial content information loss at elevated weights. To address these issues, we integrate the FAGS module to enhance the interaction among patches. The effectiveness of this enhancement is displayed in Row 4 of Fig. \ref{fig:ablation_quali}, where it leads to a more unified stylization of the left and right parts of the woods in the Watercolor results, while also preserving the color details of the source content. The increase in all metrics in Row 4 of Table \ref{tab:ablation_quanti}, compared to the baseline and method (+SWC w/. More Style), further demonstrates the effectiveness of the FAGS strategy in improving the style control loss.



\subsubsection{Effect of the Pre-Shape Self-correlation Consistency Loss}
\label{sec:Effect of the Pre-Shape Self-correlation Consistency Loss}

The implementation of the SWC and FAGS strategies significantly addresses the issue of style inconsistency and enhances the preservation of color information in the content. However, these strategies can affect the preservation of the contour shapes of the content. For instance, in the Cubism stylization displayed in Row 4, the base of the sculpture is incompletely rendered.

To better balance content preservation with the desired degree of stylization, we introduce the Pre-Shape self-correlation consistency loss. This adjustment leads to optimal visual results and achieves a balance in metrics, as evidenced in Row 5 of Fig. \ref{fig:ablation_quali} and Table \ref{tab:ablation_quanti}. The application of our refined style and content control loss clearly demonstrates improvements in visual quality, effectively resolving previous issues while enhancing overall image fidelity. 



\subsubsection{Hyperparameters Setting}
\label{sec:Hyperparameters setting}

\begin{figure}[htbp]
  \begin{center}
  \includegraphics[width=1\linewidth]{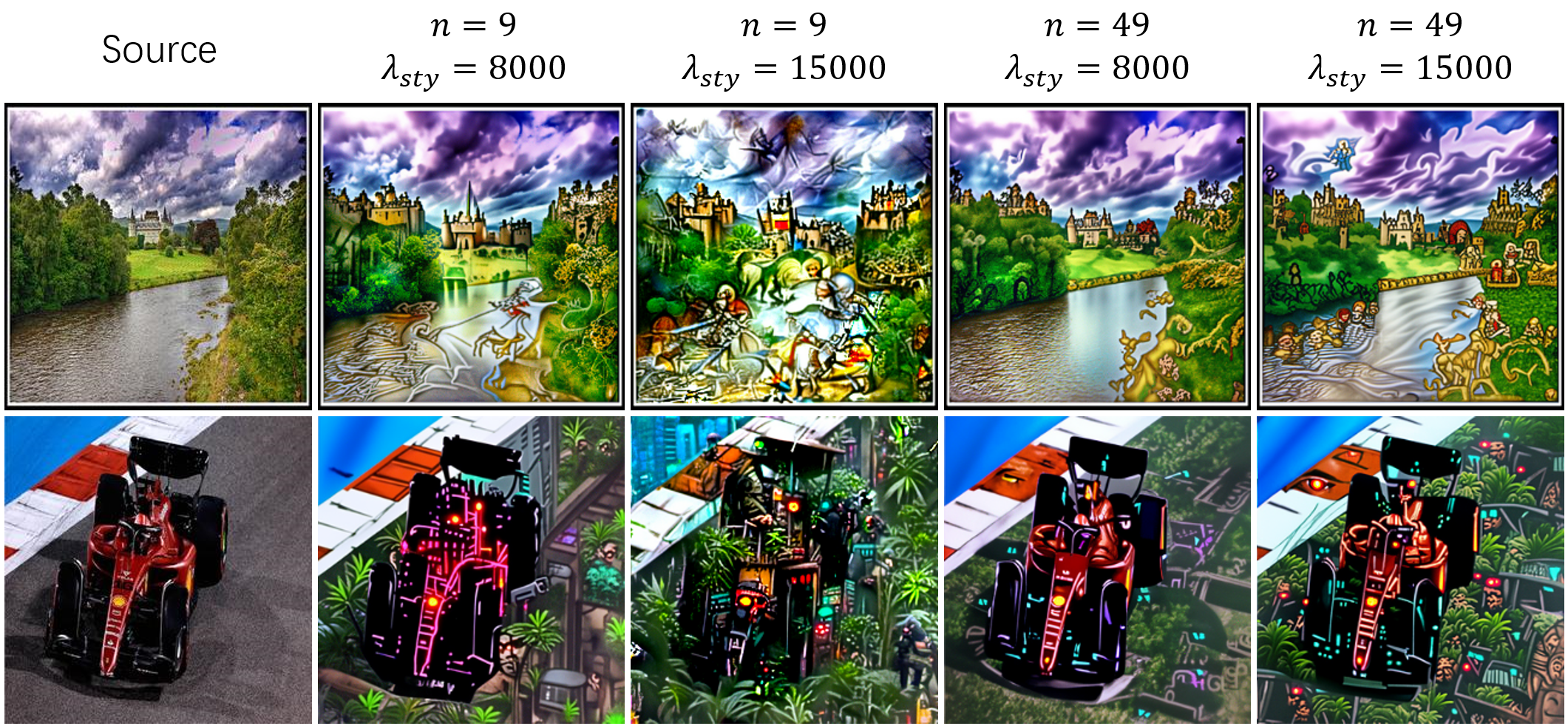}
  \end{center}\caption{The ablation study of hyperparameter $n$. The results in Row 1 correspond to the "Medieval Fantasy" style prompt, while results in Row 2 correspond to the "Jungle Cyberpunk".} 
  \label{fig:ablation_npatch}
  \end{figure}

Our method requires setting a total of 7 hyperparameters, including the number of patches $ n $ derived from SWC and the weights for $L_{pc}$, $L_{pd}$, $L_{psc}$, $L_{ZeCon}$, $L_{VGG}$ and $L_{MSE}$. 

The impact of the patch number $n$ on stylization outcomes is demonstrated in Fig. \ref{fig:ablation_npatch}. When only 9 patches are cropped, the weight of the overall style control loss significantly influences image quality. For instance, the Medieval stylization in Column B appears cluttered and loses the contour definition of the original content. Despite applying lower weight settings, content loss still occurs, as evidenced in Column A where parts of the racing car's cabin contour are missing.

Conversely, with $n=49$, the smaller patch size facilitates a more detailed preservation of content. Adjusting the overall weight of the style control loss in this setting influences the intensity of the stylization but does not substantially affect content preservation, allowing for a more controllable approach to achieving the desired stylistic effects without compromising the integrity of the original image.



\begin{figure}[htbp]
  \begin{center}
  \includegraphics[width=1\linewidth]{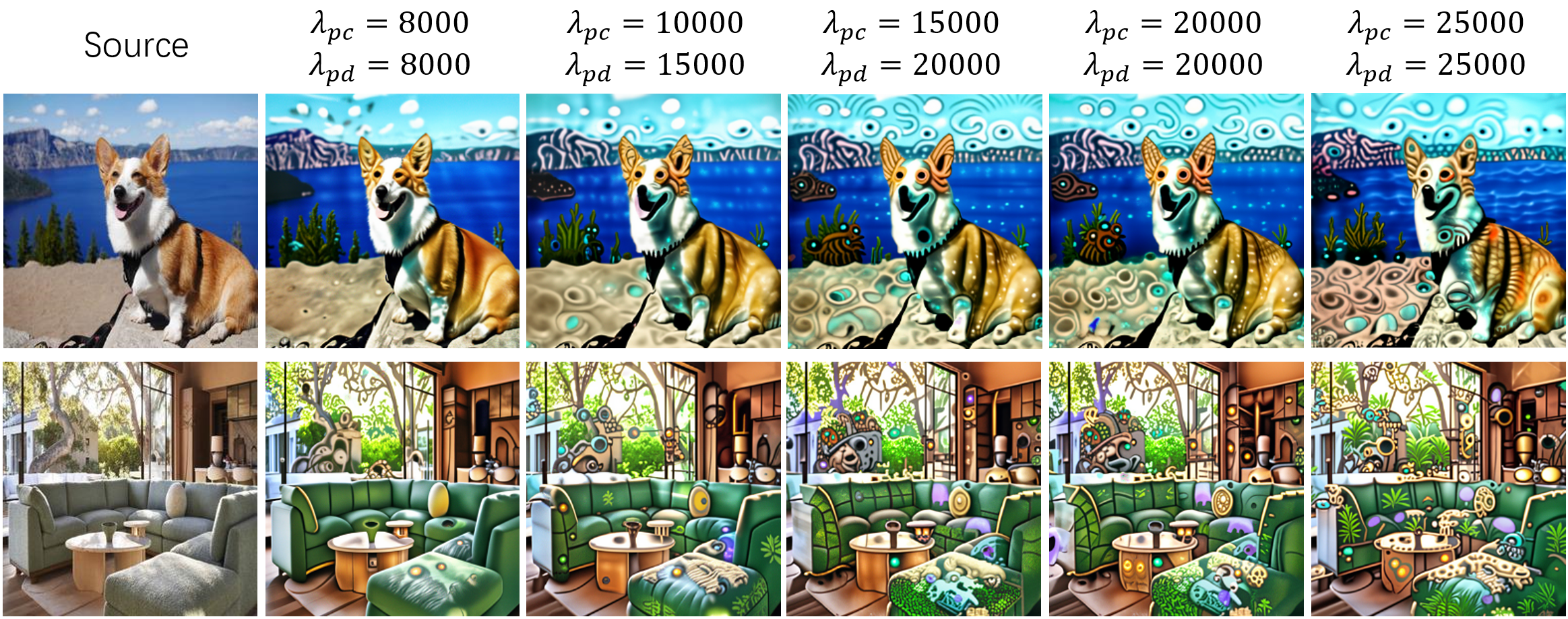}
  \end{center}\caption{The ablation study of hyperparameters $\lambda_{pc}$ and $\lambda_{pd}$. The results in Row 1 correspond to the "Deep Sea Primitive Tribal" style prompt, while results in Row 2 correspond to the "Jungle Steampunk".} 
  \label{fig:ablation_weitghts}
  \end{figure}

As depicted in Fig. \ref{fig:ablation_weitghts}, setting the weights $\lambda_{pc}$ and $\lambda_{pd}$ of $L_{pc}$ and $L_{pd}$ below 15,000 may result in insufficient stylization. Through extensive testing, we have determined that a weight range from 15,000 to 25,000 is optimal for achieving desirable stylization outcomes across various source images and style prompts. Consequently, we have established $\lambda_{pc}$ and $\lambda_{pd}$ at 20,000 as the standard settings for our method.


\begin{table}[htbp]
  \centering
  \caption{The default settings for all hyperparameters of our proposed FAGStyle.} 
  \begin{tabular}{@{}c|ccccccc@{}}
  \toprule
  Hyperparameters & $n$  & $\lambda_{pc}$ & $\lambda_{pd}$ & $\lambda_{ps}$ & $\lambda_{z}$ & $\lambda_{v}$ & $\lambda_{m}$ \\ \midrule
  Value           & 49 & 20000                          & 20000                          & 1000                           & 1000                          & 1000                          & 100                           \\ \bottomrule
  \end{tabular}
  \label{tab:hyperparameters}
  \end{table}


For the content control loss, we align our hyperparameter settings with those used in ZeCon \cite{yang2023zero}, configuring the weights $\lambda_{z}$, $\lambda_{v}$, and $\lambda_{m}$ for $L_{ZeCon}$, $L_{VGG}$, and $L_{MSE}$ at 1000, 1000, and 100, respectively. To ensure uniformity across our method, we also set $\lambda_{ps}$ for $L_{psc}$ at 1000. The default settings for all hyperparameters of our approach are detailed in Table \ref{tab:hyperparameters}.

\section{Conclusion}
\label{sec:Conclusion}


In this work, we introduce FAGStyle, a novel zero-shot text-guided diffusion image style transfer method. We enhance the interaction of information between patches by integrating the SWC technique and FAGS into our style control loss, thus improving the consistency of stylization. Additionally, we introduce a Pre-Shape self-correlation consistency loss to maintain content consistency. Extensive experimental results show that our proposed FAGStyle method outperforms existing state-of-the-art approaches in achieving superior results on both imagined and common styles. However, our method currently faces challenges with the slow construction speed of the geodesic surface. Future work will focus on accelerating the stylization process and integrating the FAGS module into a diffusion model framework that uses cross-attention to merge style information, further enhancing the efficiency and effectiveness of our style transfer approach.

\section*{Acknowledgments}
This research is sponsored by National Natural Science Foundation of China (Grant No. 52273228), Key Research Project of Zhejiang Laboratory (No. 2021PE0AC02), Key Program of Science and Technology of Yunnan Province (202302AB080022), the Project of Key Laboratory of Silicate Cultural Relics Conservation (Shanghai University), Ministry of Education (No. SCRC2023ZZ07TS).

\bibliographystyle{IEEEtran}
\bibliography{style_transfer.bib}

\end{document}